\documentclass{bmvc2k}

\title{Hybrid Video Anomaly Detection for\\Autonomous Driving}

\addauthor{Daniel Bogdoll\textsuperscript{\textasteriskcentered}}{bogdoll@fzi.de}{1,2}
\addauthor{Jan Imhof\textsuperscript{\textasteriskcentered}}{uwxdb@student.kit.edu}{1}
\addauthor{Tim Joseph}{joseph@fzi.de}{2}
\addauthor{Svetlana Pavlitska}{pavlitska@fzi.de}{1,2}
\addauthor{J. Marius Zöllner}{zoellner@fzi.de}{1,2}

\addinstitution{
Karlsruhe Institute of Technology (KIT)\\
Karlsruhe, Germany
}
\addinstitution{
FZI Research Center for Information Technology\\
Karlsruhe, Germany
}

\runninghead{D. Bogdoll et al.}{Hybrid Video Anomaly Detection for AD}

\usepackage{graphicx}
\usepackage{wrapfig}
\usepackage{booktabs}
\usepackage{amssymb}
\usepackage{booktabs}

\usepackage{pifont}
\newcommand{\cmark}{\ding{51}}%
\newcommand{\xmark}{\ding{53}}%

\begin{document}

\maketitle

\def\thefootnote{\textsuperscript{\textasteriskcentered}}\footnotetext{These authors contributed equally}\def\thefootnote{\arabic{footnote}}
\begin{abstract}
In autonomous driving, the most challenging scenarios can only be detected within their temporal context. Most video anomaly detection approaches focus either on surveillance or traffic accidents, which are only a subfield of autonomous driving. We present HF$^2$-VAD$_{AD}$, a variation of the HF$^2$-VAD surveillance video anomaly detection method for autonomous driving. We learn a representation of normality from a vehicle's ego perspective and evaluate pixel-wise anomaly detections in rare and critical scenarios.
\end{abstract}

\section{Introduction}
\label{sec:intro}

In autonomous driving, the detection of anomalies is crucial for ensuring safety and reliability. Video anomaly detection (VAD) focuses on events in video data that deviate from an expected normality. In autonomous driving, the challenges are complicated by factors such as camera movements, ever-changing backgrounds, and rapid changes in vehicle speed.

Many different types of anomalies exist~\cite{breitenstein_sys,heidecker_appl,Bogdoll_description}, with many approaches trying to detect them~\cite{bogdoll_anomaly_2022}. Reconstruction-based methods focus on learning a representation of normality and labeling deviations from this norm~\cite{hasan_learning_2016, luo_remembering_2017}. Predictive approaches detect anomalies when future predictions deviate significantly from observations~\cite{liu_future_2018, yu_cloze_2020}. However, in autonomous driving, most of the anomaly detection methods that are based on raw sensory data focus on atypical objects rather than atypical behaviors. Temporal anomalies are often detected in object-based trajectory data, assuming perfect perception~\cite{wiederer2021anomaly,jiao2023learning, breuer_quo_2021, group_interpretable_2023}. The most complex temporal anomalies are \textit{anomalous scenarios}, which pose a high risk of collision~\cite{breitenstein_sys}.

\begin{figure}[h!]
  \centering
  \includegraphics[width=1\textwidth, keepaspectratio]{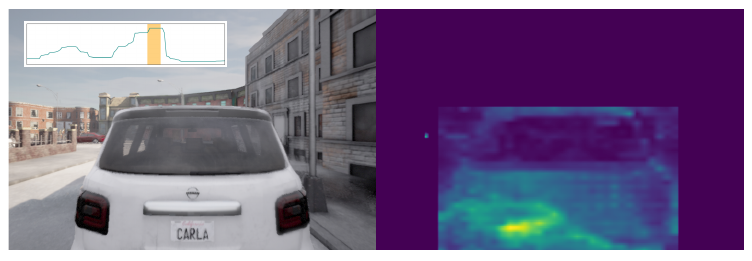}
\caption{Exemplary anomaly detection with HF$^2$-VAD$_{AD}$. The vehicle in front performs a sudden braking maneuver, as highlighted in yellow in the top-left graph. The graph shows the anomaly scores per frame, peaking while the vehicle performs the maneuver. The right frame shows pixel-wise anomaly scores for all instances of the class \textit{vehicle}, computing anomaly scores only within class-specific bounding boxes predicted by an object detector.}
  \label{fig:graph_beginning}
\end{figure}

Until recently, mostly accident-focused benchmarks were available to evaluate temporal anomalies in the context of autonomous driving~\cite{fangVisionBasedTrafficAccident2024}. Given a rise in datasets with novel dynamic anomalies~\cite{wang_deepaccident_2023,gong_sdac_2024,bogdoll2024anovox}, we are interested in the applicability of video anomaly detection methods for autonomous driving. Among recent SotA methods, we choose HF$^2$-VAD as the representative model~\cite{liu_hybrid_2021} and adapt it to the field of autonomous driving. For this purpose, we train a HF$^2$-VAD$_{AD}$ model on simulated data representing normality and evaluate it on the AnoVox benchmark~\cite{bogdoll2024anovox} with \textit{anomalous scenarios} in the form of sudden braking maneuvers of a lead vehicle. We also introduce a pixel-wise evaluation to localize anomalies.

\section{Related Work}
\label{sec:sota}

There are three typical camera perspectives for anomaly detection methods in the traffic domain. \textit{Birds-Eye-View (BEV)} and \textit{surveillance} perspectives come from external data providers, while an ego or \textit{dashcam} view can be provided by a vehicle itself~\cite{fangVisionBasedTrafficAccident2024}. Numerous works have focused on specific scenarios, including the detection of atypical activity in urban surveillance, the analysis of irregularities in pedestrian dynamics, and the monitoring of abnormal behavior patterns in crowds~\cite{chandola_anomaly_2009, popoola_video-based_2012, nayak_comprehensive_2021, ramachandra_survey_2020, tran_anomaly_2022, santhosh_anomaly_2020, aldayri_taxonomy_2022, patil_survey_2016, chandrakala_anomaly_2023} as well as street scenes recorded by city cameras that evaluate vehicle behavior from an outside perspective~\cite{kumaran_anomaly_2021, djenouri_survey_2019}.

Especially the VAD subfields traffic anomaly detection~\cite{liang_tad,zhaoGoodPracticesStrong} and road accident detection~\cite{fangAbductiveEgoViewAccident2024} have gained much traction. However, only a few methods specifically analyze behavioral anomalies from the perspective of a vehicle's onboard cameras. For our comparison in Table~\ref{tab:sota}, we have also excluded frame-wise~\cite{zhouSpatioTemporalFeatureEncoding2022,hareshAnomalyDetectionDashcam2020,lagraaRealTimeAttackDetection2019,aivatoglouEVENTRealtimeVideo2023} or supervised~\cite{chanAnticipatingAccidentsDashcam2017,kimCrashNotCrash2019,leAttentionRCNNAccident2020,hajriVisionTransformersRoad2022} methods as well as such that focus on domain-level anomalies~\cite{stoccoMisbehaviourPredictionAutonomous2020,hussainDeepGuardFrameworkSafeguarding2021,grewalPredictingSafetyMisbehaviours2024,bai_recognizing_2018}.

In general video anomaly detection, hybrid methods that combine both reconstruction and prediction methods have emerged as the current state-of-the-art. These methods aim to utilize the strengths of both approaches by leveraging temporal dynamics and reconstruction capabilities to improve anomaly detection~\cite{ye_anopcn_2019, tang_integrating_2020, liu_hybrid_2021, liu_spatio-temporal_2022, li_dual-branch_2022, huang_novel_2023}. Among these, HF$^2$-VAD by Liu et al.~\cite{liu_hybrid_2021}, who built upon the work of Park et al.~\cite{parkLearningMemoryGuidedNormality2020}, is commonly used in the literature as a benchmark for the state of the art in video anomaly detection~\cite{ristea_self-distilled_2023} and often serves as the foundation for further advancements~\cite{wan_pose-motion_2023}. It detects unusual events that deviate from normal patterns for surveillance tasks.

The idea of predicting future frames for the purpose of anomaly detection became popular around 2019, where it was demonstrated both for surveillance tasks~\cite{liu_future_2018} and autonomous driving~\cite{bolte_towards,yaoUnsupervisedTrafficAccident2019}. Liu et al. combined future frame predictions with an optical flow~\cite{liu_future_2018} in \textbf{AnoPred}, and \textbf{CC-Detector} by Bolte et al. focused on semantically distinguishing relevant classes from the rest of an image~\cite{bolte_towards}. Fang et al. proposed self-supervised consistency learning for traffic accident detection~(\textbf{SSC-TAD})~\cite{fangTrafficAccidentDetection2022}, where the key idea is to detect deviations of appearance, motion, and context relations. For this, they also utilize an explicit graph representation of the scenes. Future object localization (\textbf{FOL}) by Yao et al., on the other hand, focused on predicting only the bounding boxes of relevant classes in order to compensate for the motion of the ego vehicle, which makes the prediction of whole frames more complex~\cite{yaoUnsupervisedTrafficAccident2019}. Later, Fang et al. built upon their work but removed the need for an odometry estimation for the ego vehicle. Recently, Yao et al. extended their own work and also provided updated \textbf{AnoPred$_{Mask}$} version of Liu's approach by adding masks for relevant objects, similar to Bolte's approach~\cite{yaoDoTAUnsupervisedDetection2023}. Finally, the new DeepAccident dataset by Wang et al. proposed \textbf{AccPred}, an accident detection method based on future trajectories and their distances to each other~\cite{wang_deepaccident_2023}. As shown in Table~\ref{tab:sota}, most of the methods require many additional sub-methods to function, making them rather complex and only recently began to work for high-resolution input images.

\begin{table}[t]
\centering
\resizebox{\textwidth}{!}{%
\begin{tabular}{@{}lccccccccccr@{}}
\toprule
\textbf{Method}        & \textbf{Year} & \textbf{High Resolution}       & \multicolumn{8}{c}{\textbf{Required Sub-Methods}}                                                                                                                                                                             & \textbf{Anomaly Detection} \\ \midrule
                       &               &                                & \textbf{ROI}              & \textbf{(BEV) Mask}  & \textbf{Bbox}             & \textbf{Flow}             & \textbf{Odometry}         & \textbf{Tracking}         & \textbf{Motion Pred.}      & \textbf{Graph}            &                            \\
\textbf{AnoPred~\cite{liu_future_2018,yaoUnsupervisedTrafficAccident2019,yaoDoTAUnsupervisedDetection2023}}       & 2019          & ---                            & \textbf{}                 &  \xmark & \textbf{}                 & \textbf{}                 & \textbf{}                 & \textbf{}                 & \textbf{}                 & \textbf{}                 & \textbf{Learned Normality} \\
\textbf{CC-Detector~\cite{bolte_towards}}         & 2019          & ---                            &  \xmark &  \xmark & \textbf{}                 & \textbf{}                 & \textbf{}                 & \textbf{}                 & \textbf{}                 & \textbf{}                 & \textbf{Learned Normality} \\
\textbf{SSC-TAD~\cite{fangTrafficAccidentDetection2022}}          & 2022          & ---                   & \textbf{}                 & \textbf{}                 &  \xmark &  \xmark & \textbf{}                 &  \xmark & \textbf{}                 &  \xmark & \textbf{Learned Normality} \\
\textbf{FOL~\cite{yaoDoTAUnsupervisedDetection2023,yaoUnsupervisedTrafficAccident2019}}           & 2023          & \textbf{\cmark} & \textbf{}                 & \textbf{}                 &  \xmark &  \xmark &  \xmark &  \xmark & \textbf{}                 & \textbf{}                 & \textbf{Learned Normality} \\
\textbf{AnoPred$_{Mask}$~\cite{liu_future_2018,yaoDoTAUnsupervisedDetection2023}} & 2023          & ---                            & \textbf{}                 & \textbf{}                 & \textbf{}                 &  \xmark & \textbf{}                 & \textbf{}                 & \textbf{}                 & \textbf{}                 & \textbf{Learned Normality} \\

\textbf{AccPred~\cite{wang_deepaccident_2023}}  & 2023          & \textbf{\cmark} & \textbf{}                 &  \xmark &  \xmark & \textbf{}                 & \textbf{}                 & \textbf{}                 &  \xmark & \textbf{}                 & Metric-based               \\ \midrule
\textbf{HF$^2$-VAD$_{AD}$~\cite{liu_hybrid_2021}}   & 2024          & \textbf{\cmark} & \textbf{}                 & \textbf{}                 &  \xmark &  \xmark & \textbf{}                 & \textbf{}                 & \textbf{}                 & \textbf{}                 & \textbf{Learned Normality} \\ \bottomrule
\vspace{2pt}
\end{tabular}%
}
\caption{Behavioral, pixel-wise anomaly detection methods for vehicle camera data.}
\label{tab:sota}
\end{table}

\section{Methodology}
\label{sec:method}

As shown in Section~\ref{sec:sota} and Table~\ref{tab:sota}, existing video anomaly detection methods for autonomous driving are either overly complex or incapable of dealing with high-resolution input data. Based on HF$^2$-VAD~\cite{liu_hybrid_2021}, our adaption HF$^2$-VAD$_{AD}$ is capable of utilizing high-resolution video streams, learns a representation of normality in the domain of autonomous driving and only requires two sub-methods. Thus, it is the least complex method for video anomaly detection in the field of autonomous driving for high-resolution input. 


\begin{figure}[h!]
    \centering
     \resizebox{1\textwidth}{!}{
        \includegraphics{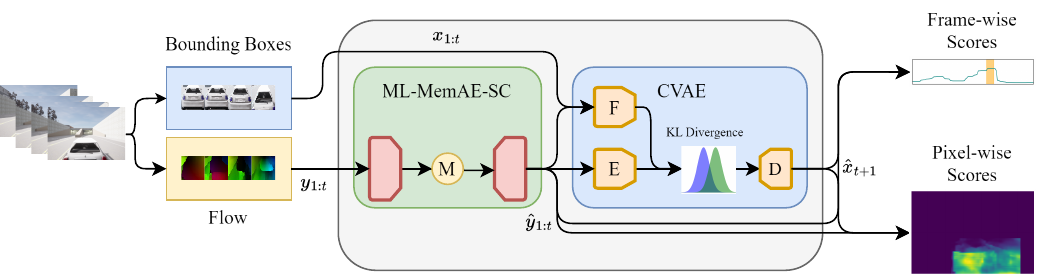}
     }
     
\caption{HF$^2$-VAD$_{AD}$: Adaptation and extension of HF$^2$-VAD for autonomous driving. Optical flows ${y}_{1:t}$ and bounding boxes ${x}_{1:t}$ for relevant objects are generated for each input image. ML-MemAE-SC reconstructs the optical flows $\hat{y}_{1:t}$ with memory-modules $M$ to better reconstruct only normal patterns. The CVAE predicts a future frame $\hat{x}_{t+1}$. On this basis, image-wise and localized pixel-wise anomaly scores are generated. Adapted from~\cite{liu_hybrid_2021}.}
  \label{fig:paper-process}
\end{figure}

\textbf{HF$^2$-VAD.} Here, we provide a brief overview of the hybrid framework HF$^2$-VAD~\cite{liu_hybrid_2021} we build upon and refer to the original publication for further details. The framework combines Multi-Level Memory modules in an Autoencoder with Skip Connections (ML-MemAE-SC) for optical flow reconstructions and a Conditional Variational Autoencoder (CVAE) for future frame predictions. The framework is trained on normal data and uses discrepancies in the reconstructions and predictions to identify anomalies. Experiments were conducted with well-established surveillance datasets.

The \textit{ML-MemAE-SC} uses an autoencoder structure with multiple memory modules at different levels. This structure improves the model's ability to learn and retain typical data patterns from the training set and distinguish them from anomalies. The integration of skip connections between the encoder and decoder helps to preserve important data that might otherwise be lost during the encoding process. This part focuses on the accurate reconstruction of normal data. The training of the ML-MemAE-SC includes a composite loss function that summarizes reconstruction and entropy losses. The reconstruction loss is calculated based on the Euclidean distance.

The \textit{CVAE} plays a central role in predicting future images by using conditioned historical data to simulate probable future scenarios. It determines the latent variable \( z_{t+1} \) by considering both the current image \( x_t \) and the optical flow \( y_{t,t+1} \) between successive images. It works on the premise that a latent variable \( z \), representing the probabilistic state of the following image, can be derived from the immediate past to subsequently influence the generation of \(\hat{x}_{t+1} \). It uses the encoded optical flow and the current image to estimate the posterior distribution \( p(z_{t+1} \mid x_t, y_{t,t+1}) \). A sample from this distribution, which represents the probabilistic state of the next image, is then drawn for image generation. During training, this latent variable \( z \) is combined with the conditions derived from the encoder \( E \) and processed by the decoder \( D \) to create the future image \( \hat{x}_{t+1} \).

At test time, anomaly detection is performed frame-wise. The frame-wise anomaly score \(S_{f}\) is a fusion of the flow reconstruction error \(S_{r}\) and
the error in the prediction of future images \(S_{p}\), standardized by their respective means \(\mu\) and standard deviations \(\sigma\) from the training samples:

\begin{equation}
S_{f} = w_{r} \cdot \frac{S_{r} - \mu_{r}}{\sigma_{r}} + w_{p} \cdot \frac{S_{p} - \mu_{p}}{\sigma_{p}}
\end{equation}




\textbf{HF$^2$-VAD$_{AD}$.} As shown in Figure~\ref{fig:paper-process}, we first adapt HF$^2$-VAD to autonomous driving by generating dense, pixel-wise anomaly scores for the whole frame rather than focusing on the content of unlocalized, detected bounding boxes alone, as can be seen in Figure~\ref{fig:graph_beginning}. We learn flow reconstructions and flow-guided frame prediction from the perspective of an ego-vehicle rather than from the perspective of a static surveillance camera. Our training dataset was collected with the CARLA simulation engine~\cite{dosovitskiy_carla_2017}. Rather than collecting data with the default autopilot, we employ Roach~\cite{zhang_end--end_2021} as the driving agent.

To reflect the complexity of real-world traffic, the training dataset features four different cities, each with its own weather and traffic dynamics. The training data includes 38 scenarios with over 30,000 high-resolution frames with a resolution of 600 x 800 pixels. Crucially, the training dataset contains only normal driving scenes so that the model can internalize a baseline of normal traffic behavior, which is crucial for detecting deviations that indicate anomalies in real-world driving situations. Other normally behaving agents, including lead vehicles, are included in the training data.

Flows are generated by FlowNet2.0~\cite{ilg_flownet_2016} to ensure that both static and motion-based features contribute to the training of the model. Bounding boxes are detected with an off-the-shelf object detection model~\cite{carla_carla_2022} trained on CARLA data. We first train the ML-MemAE-SC with the extracted flows from the training dataset, then train the CVAE model with the reconstructed flows from the ML-MemAE-SC and the corresponding frames, and finally fine-tune the whole framework. During fine-tuning, the weights of the two models are adjusted together to improve their coordination and the overall accuracy of anomaly detection.

We train the model specifically on foreground objects, i.e., \textit{vehicles}, extracted from these images with bounding boxes, and construct spatiotemporal cubes (STC) for each object, where the dimensions of these cubes are standardized to 32x32. The use of bounding boxes for each vehicle in the scene allows the model to focus specifically on regions within the image where temporal anomalies in the form of unexpected maneuvers can occur.

\begin{figure}[t!]
  \centering
  \includegraphics[width=1\textwidth, keepaspectratio]{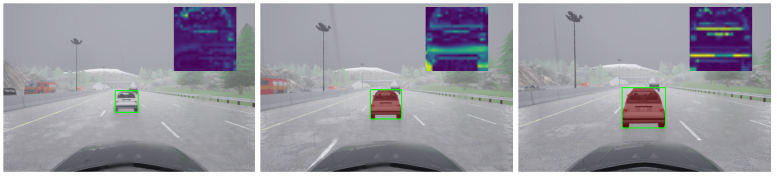}
\caption{\textbf{Sequential Anomaly Detection:} The first image shows the normal driving mode of the lead vehicle, while the last two images show an anomaly with the pixel-wise ground truth overlayed in red. Other regularly driving traffic participants are also present. As visible in the anomaly maps, HF$^2$-VAD$_{AD}$ successfully detects the unknown maneuver.}
  \label{fig:braking}
\end{figure}

In addition to frame-wise anomaly detection, we have integrated a pixel-wise anomaly detection. The pixel-wise anomaly score, denoted as \( S_{\text{pixel}} \), is computed by combining the MSE values from optical flow and image prediction, both of which are normalized to enhance the reliability of the anomaly detection process:

\[
S_{\text{pixel}} = w_{\text{r,p}} \cdot m_{\text{r}} + w_{\text{p,p}} \cdot m_{\text{p}}
\]

Here, \( m_{\text{r}} \) and \( m_{\text{p}} \) represent the aggregated robustly scaled MSE values for optical flow and image prediction, respectively. To enhance the robustness and comparability of these MSE values across different datasets, a robust scaling technique is applied. This scaling is important as it mitigates the influence of outliers and non-standard data distributions, thereby improving the reliability of the anomaly detection process. The robust scaling of each MSE value x is defined as follows:

\[
x_{\text{scaled}} = \frac{x - \tilde{x}}{\text{IQR}(X)}
\]

Where \( x \) is the original MSE value, \( \tilde{x} \) is the median of the MSE values across the dataset, and \( \text{IQR}(X) \) is the interquartile range, defined as the difference between the 75th and 25th percentiles (\( Q_{75}(X) \) and \( Q_{25}(X) \)). After robust scaling, the mean value is calculated for each pixel across the RGB channels:

\[
m_{\text{aggregated}} = \frac{1}{N} \sum_{i=1}^{N} x_{\text{scaled}, i}
\]

Where \( N \) is the number of RGB channels. The final pixel-wise anomaly score \( S_{\text{pixel}} \) is computed by combining the aggregated MSE values from optical flow and image prediction using the weights \( w_{\text{r,p}} \) and \( w_{\text{p,p}} \).

The resulting pixel-wise scores are then used to localize anomalies within the image. A score of 0 is set for all pixels outside the bounding boxes, as shown in Figure~\ref{fig:graph_beginning}. In contrast, the frame-wise approach of the original HF$^2$-VAD uses simple summation and normalization based on the training statistics. Specifically, the sum of MSE losses across all dimensions is calculated for both optical flow and image predictions, and these summed values are then normalized using the mean and standard deviation from the training data. In the presence of detections, the authors added up all $32 \times 32$ anomaly patches and chose the largest value as the frame-wise anomaly score. We report these frame-wise results as AUROC$^{\diamond}$ in Section~\ref{sec:experiments}. 

In contrast, our approach calculates a pixel-wise anomaly score by normalizing and aggregating mean-squared errors of optical flow and frame predictions, then scaling these scores to bounding boxes within each frame. Pixel-wise scores allow for a localization of anomalies for a more granular analysis, compared to a single aggregated score per frame.
\section{Evaluation}
\label{sec:experiments}

\begin{figure}[t!]
  \centering
  \includegraphics[width=0.9\textwidth, keepaspectratio]{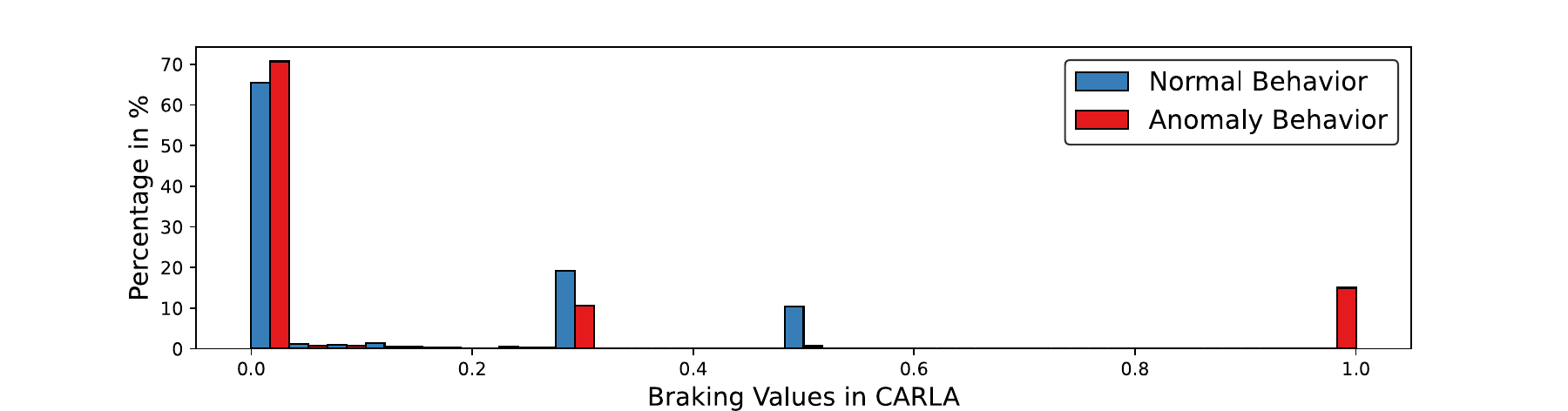}
\caption{Distribution of braking behaviors of other traffic participants in training (blue) and test (red) data. High-intensity sudden braking scenarios, labeled as anomalous, only occur in the evaluation dataset.}
  \label{fig:histogram}
\end{figure}

For the evaluation, we used the AnoVox benchmark~\cite{bogdoll2024anovox} with \textit{anomalous scenarios} in the form of sudden braking maneuvers performed by lead vehicles, as visible in Figure~\ref{fig:braking}. These are common and dangerous situations in road traffic~\cite{kim_crash_2019} and are not included in our training data, as Figure~\ref{fig:histogram} shows. In the 13 evaluation scenarios, the ego-vehicle follows a lead vehicle along a longer route under different weather and traffic conditions. At certain intervals, the lead vehicle performs a sudden braking maneuver. Regularly behaving traffic participants are also present in the evaluation dataset. Based on the available pixel-wise ground truth annotations of sudden braking scenarios in AnoVox, as shown in Figure~\ref{fig:braking}, we also generated bounding-box and frame-wise annotations for the evaluation. In AnoVox, the anomalies are annotated starting with the unexpected sudden braking of the leading vehicle and ending with its complete stop.

\begin{table}[h!]
\centering
\renewcommand{\arraystretch}{1.3} 
\resizebox{0.9\columnwidth}{!}{
\begin{tabular}{lcccccc}
\toprule
\textbf{Model} & \textbf{Domain} & \textbf{Weather} & \textbf{Finetuned} & \textbf{AUROC$^{\diamond}$} $\uparrow$ & \textbf{FPR$_{95}$ $\downarrow$} & \textbf{IoU} $\uparrow$ \\ \hline
\multicolumn{7}{c}{\textit{Bounding Boxes: Detection}} \\ \hline
HF$^2$-VAD$_{AD}$ & All & All & \checkmark & 66.98 &  \textbf{2.58} & 48.10 \\
HF$^2$-VAD$_{AD}$ & City & Sunshine  & \checkmark & 68.04 & 2.48 & 48.09 \\
HF$^2$-VAD$_{AD}$ & City & Rain & \checkmark & \underline{69.74} & \underline{2.68} & \underline{51.30} \\
HF$^2$-VAD$_{AD}$ & Highway & Sunshine  & \checkmark & 57.22 & 3.15 & \textbf{60.71}  \\
HF$^2$-VAD$_{AD}$ & Highway & Rain & \checkmark & \textbf{76.60} & \textbf{1.34} & 41.09 \\
\multicolumn{7}{c}{\textit{Bounding Boxes: Ground Truth}} \\ \hline
ML-MemAE-SC & All & All & --- & 67.49 & 7.80 &--- \\
HF$^2$-VAD$_{AD}$ & All & All & --- & \underline{81.01} & 3.72 &--- \\
HF$^2$-VAD$_{AD}$ & All & All & \checkmark & 79.09 & \textbf{2.83} &--- \\
HF$^2$-VAD$_{AD}$ & City & Sunshine & \checkmark & 80.74 & 3.10 & --- \\
HF$^2$-VAD$_{AD}$ & City & Rain & \checkmark & 74.08 & \underline{3.05} & --- \\
HF$^2$-VAD$_{AD}$ & Highway & Sunshine & \checkmark & 74.76 & 3.19 & ---\\
HF$^2$-VAD$_{AD}$ & Highway & Rain & \checkmark & \textbf{91.74} & 99.32 & ---\\
\bottomrule
\vspace{2pt}
\end{tabular}
}
\caption{Evaluation and ablation studies of HF$^2$-VAD$_{AD}$ and ML-MemAE-SC under different scenarios, with \textbf{best} and \underline{second-best} results highlighted.}
\label{tab:quantitative_results}
\end{table}

In order to evaluate the performance of HF$^2$-VAD$_{AD}$, we provide both quantitative and qualitative insights. Two metrics are used for evaluation: the already used Area Under the Receiver Operation Characteristic (AUROC$^{\diamond}$) for frame-wise evaluations and the False Positive Rate at 95\% True Positive Rate (FPR$_{95}$) for pixel-wise evaluation. The AUROC metric, widely used in the VAD literature~\cite{gong_memorizing_2019, liu_hybrid_2021, lu_abnormal_2013, mahadevan_anomaly_2010}, serves as a measure of VAD accuracy by varying the threshold for anomaly assessment, with higher AUROC values indicating better detection performance. Adding the FPR$_{95}$~\cite{henriksson_performance_2021, di_biase_pixel-wise_2021} metric provides insights into the false-positive rates if nearly all anomalies are correctly detected, with lower values indicating better detection performance. Since our detected bounding boxes do not fully match the existing ground truth, we only evaluate the FPR$_{95}$ based on the overlapping area and also provide the IoU to demonstrate the quality of the bounding boxes.

\begin{figure}[t]
    \centering
    \begin{minipage}[b]{0.49\textwidth}
        \centering
        \includegraphics[width=\textwidth]{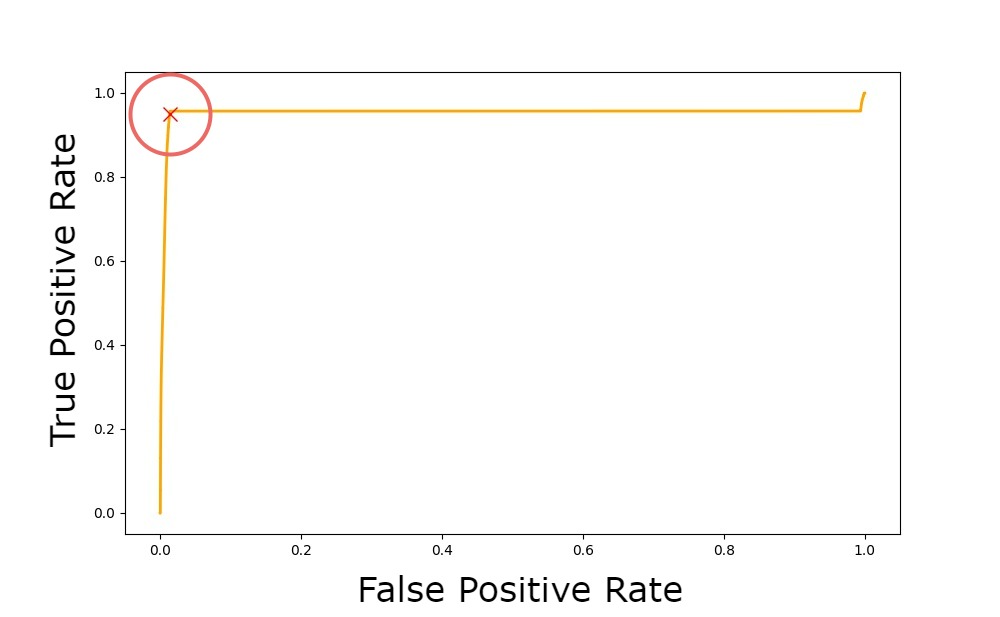} 
        \label{fig:roc1}
    \end{minipage}
    \hfill
    \begin{minipage}[b]{0.49\textwidth}
        \centering
        \includegraphics[width=\textwidth]{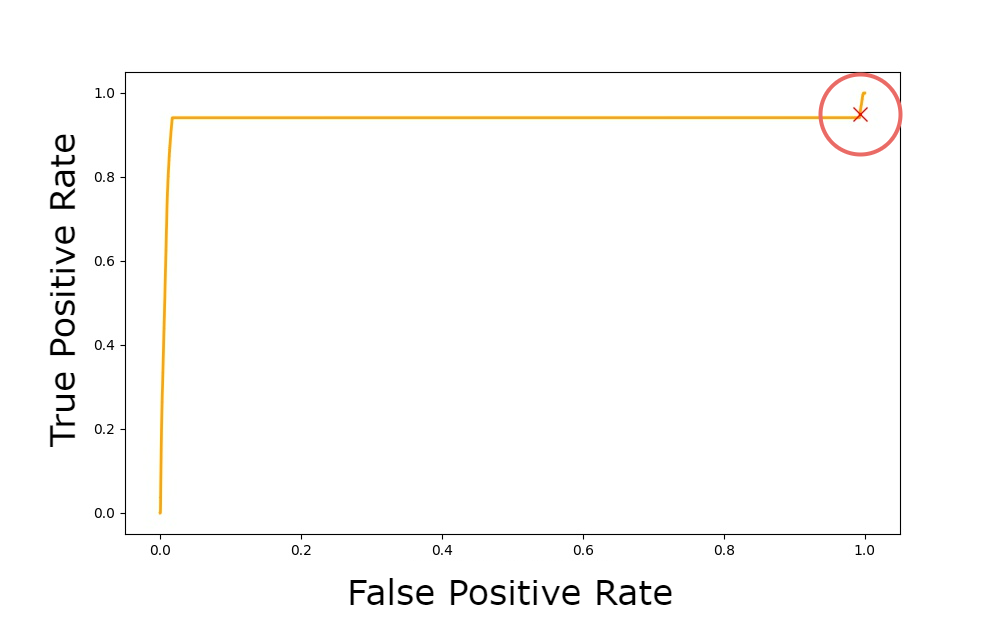} 
        \label{fig:roc2}
    \end{minipage}
    \caption{Comparison of ROC curves for the setting \textit{highway with rain}. The x marks the FPR$_{95}$ position. On the left, a low FPR$_{95}$ is achieved with predicted bounding boxes. On the right, the FPR$_{95}$ rose sharply with ground truth bounding boxes.}
    \label{fig:roc_comparison}
\end{figure}

\subsection{Quantitative Results}

As shown in Table~\ref{tab:quantitative_results}, we perform various experiments with varying conditions. Comparing city and highway scenarios, we observe a strong sensitivity of the model to these environments, but no clear trend emerges. For highway settings with bad weather, we observe poor performance. As the pixel-wise anomaly scores are only calculated within bounding boxes and all other pixels are set to 0, there is generally a low false positive rate. However, if more than 5\% of the ground truth anomalies are not detected, detecting all anomalous pixels requires a threshold close to zero. This results in the entire image being detected as anomalous, causing the FPR$_{95}$ to rise sharply, as shown in Figure~\ref{fig:roc_comparison}.

\begin{table}[h]
    \centering
    \begin{tabular}{cccc}
        \toprule
        \textbf{\(w_r\) and \(w_{r_p}\)} & \textbf{\(w_p\) and \(w_{p_p}\)} & \textbf{FPR$_{95}$} $\downarrow$ & \textbf{AUROC$^{\diamond}$} $\uparrow$ \\
        \midrule
        10 & 0.1 & \textbf{2.83} & \underline{79.09} \\
        1 & 0.1 & \underline{2.86} & 78.78 \\
        1 & 0 & \textbf{2.83} & \textbf{79.13} \\
        0 & 1 & 96.67 & 69.15 \\
        0.1 & 1 & 4.05 & 76.20 \\
        0.1 & 10 & 95.78 & 70.70 \\
        \bottomrule
        \vspace{2pt}
    \end{tabular}
    \caption{Empirical study to determine the effects of different weights for HF$^2$-VAD$_{AD}$.}
    \label{tab:weighting_results}
\end{table}

To determine optimal values for the weights, as introduced in Section~\ref{sec:method}, we perform an empirical study, as shown in Table~\ref{tab:weighting_results}. While we observe best results in the case when we completely neglect image predictions, we decided to set $w_{r}$, $w_{r_p} = 10$ and $w_{p}$, $w_{p_p} = 0.1$, as the performance metrics were too small to be significant, and we retain the advantages of both modules. The study shows that the model performs significantly worse when image prediction is weighted higher, which is in contrast to the original implementation of HF$^2$-VAD, in which the frame prediction was weighted higher than the flow reconstruction, with the weight for frame prediction set to $1$ and the weight for flow reconstruction set to $0.1$.

\textbf{Ablation Studies.} To better understand the performance of our approach, we perform three types of ablation studies, as shown in Table~\ref{tab:quantitative_results}, with the focus on the first three entries in the lower part of the table. First, we only utilize the intermediate outputs of the ML-MemAE-SC to evaluate isolated flow reconstructions. 

Second, we evaluate both bounding boxes generated from an off-the-shelf object detection model~\cite{carla_carla_2022} and ground truth bounding boxes from the evaluation dataset to isolate the effect of model failures. We observe that bounding boxes are best detected in highway settings with good weather. The IoU performance suffers especially from poor detections of distant objects. Since mostly irrelevant vehicles are missing in the detections, we see improved FPR$_{95}$ values for bounding box detections, but at the cost of reduced AUROC scores.

Lastly, we examine the fine-tuning stage during training. After fine-tuning, HF$^2$-VAD$_{AD}$ shows an improvement in precision. While the AUROC decreases slightly, the model achieves a better balance in the performance metrics with a significantly lower FPR$_{95}$.

\subsection{Qualitative Results}

\begin{figure}[t!]
  \centering
  \includegraphics[width=1\textwidth, keepaspectratio]{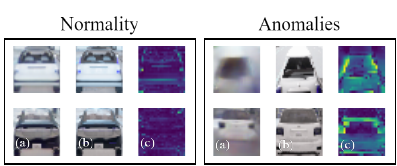}
\caption{On the left, two normal scenarios are shown. Here, the model's next-frame predictions (a) match the actual next frames (b) closely. On the right, we have shown sudden brake scenarios. Here, significant prediction differences arise, leading to high anomaly values (c).}
  \label{fig:anomaly_matrix}
\end{figure}
 
In Figure~\ref{fig:anomaly_matrix}, we show a series of comparative visualizations demonstrating the performance of HF$^2$-VAD$_{AD}$ in traffic scenarios. The model is able to predict regular behaviors very well, while atypical ones lead to poor predictions and, thus, high anomaly scores. The pixel-wise anomaly maps serve as a meaningful indicator, with more highlighted areas signaling larger prediction discrepancies. For normal driving patterns, these maps show only small deviations from the ground truth, indicating that the model can predict the subsequent images well. For anomalies, the model often predicts, as it learned, that the vehicle maintains its distance. This leads to positive anomaly detections, as the vehicle is currently anomalous and getting closer due to the braking maneuver. This can be recognized by the increased highlighting in the difference maps.

\begin{figure}[t!]
    \centering
    \includegraphics[width=1\textwidth]{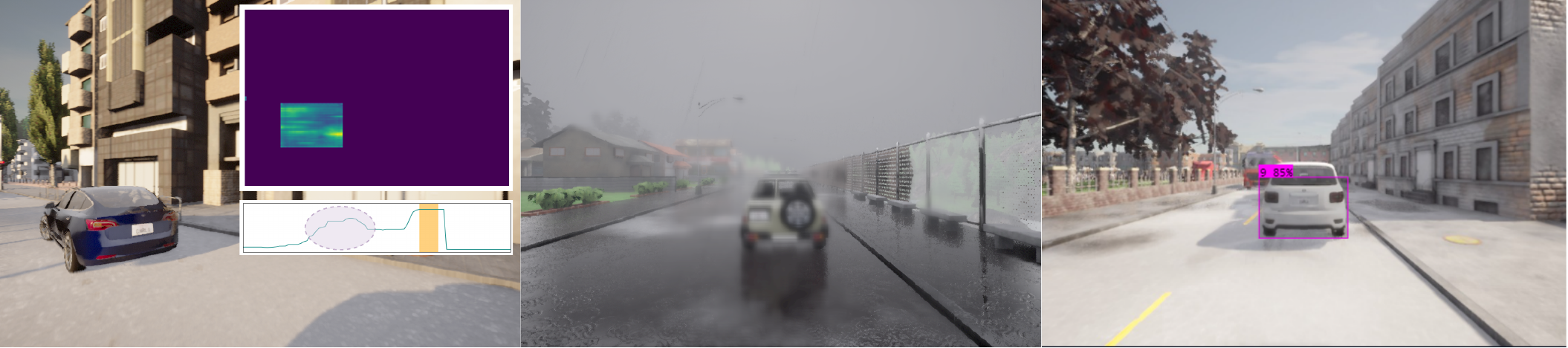}
    \caption{\textbf{Failure Cases:} Left: The model can produce high pixel-wise anomaly scores for turning vehicles. The encircled area in the frame-wise graph shows the duration of the curve. Middle: A missing bounding box detection during bad weather situations leads to no anomaly detection. Right: The bounding box only captures a part of the lead vehicle, and a truck in the back is completely missed.}
    \label{fig:failure_case}
\end{figure}

\textbf{Failure Cases.} As shown in Figure~\ref{fig:failure_case}, there are certain cases which are challenging. The first image shows how the model often incorrectly produces high anomaly scores when a lead vehicle is turning rather than going straight. 
This may be due to the model primarily learns straight-line driving patterns and the training data does not contain enough examples of lead vehicles turning directly in front of the car. The middle image shows a scene where the object detection model has problems with detecting objects in unfavorable weather conditions, such as heavy rain, so anomaly detection is not possible.

\section{Conclusion}
\label{sec:conclusion}

In summary, our research has shown that HF$^2$-VAD, a framework originally developed for detecting anomalies in surveillance systems, can be effectively transferred to autonomous driving. By extending it to dense pixel-wise anomaly maps and creating a dedicated training and test dataset, we have validated the effectiveness of HF$^2$-VAD$_{AD}$ in the vehicle context, bridging the gap between static monitoring and dynamic vehicle environments.

Our contributions further include the extension to dense anomaly detections, domain-specific weight adaptions, and dealing with multiple bounding boxes separately rather than combining them into one score, using robust scaling. In the future, we aim to further refine the framework to better deal with current failure cases and evaluate more types of anomalies to examine its performance among a wider variety of scenarios, such as wrong-way drivers.

\section*{Acknowledgment}
\label{sec:ackno}

This work results from the just better DATA project supported by the German Federal Ministry for Economic Affairs and Climate Action (BMWK), grant number 19A23003H.


\bibliography{egbib}
\end{document}